\documentclass{article}

\usepackage{microtype}
\usepackage{graphicx}
\usepackage{subfigure}
\usepackage{booktabs} 

\usepackage{hyperref}

\usepackage[accepted]{icml2024}

\usepackage{amsmath}
\usepackage{amssymb}
\usepackage{mathtools}
\usepackage{amsthm}
\usepackage{multirow}

\usepackage[capitalize,noabbrev]{cleveref}

\theoremstyle{plain}

\theoremstyle{definition}

\theoremstyle{remark}

\usepackage[textsize=tiny]{todonotes}

\icmltitlerunning{Graph neural network for ice sheet dynamics}

\begin{document}

\twocolumn[
\icmltitle{Graph Neural Networks for Emulation of Finite-Element\\Ice Dynamics in Greenland and Antarctic Ice Sheets}



\icmlsetsymbol{equal}{*}

\begin{icmlauthorlist}
\icmlauthor{Younghyun Koo}{CSE,CEE}
\icmlauthor{Maryam Rahnemoonfar}{CSE,CEE}
\end{icmlauthorlist}

\icmlaffiliation{CSE}{Department of Computer Science and Engineering, Lehigh University, Bethlehem, PA, USA}
\icmlaffiliation{CEE}{Department of Civil and Environmental Engineering, Lehigh University, Bethlehem, PA, USA}

\icmlcorrespondingauthor{Maryam Rahnemoonfar}{maryam@lehigh.edu}

\icmlkeywords{Ice-sheet and Sea-level System Model (ISSM), finite element analysis (FEA), surrogate model, graph convolutional network (GCN), graph attention network (GAT), climate change, von Mises tensile stress}

\vskip 0.3in
]



\printAffiliationsAndNotice{}  

\begin{abstract}
Although numerical models provide accurate solutions for ice sheet dynamics based on physics laws, they accompany intensified computational demands to solve partial differential equations. In recent years, convolutional neural networks (CNNs) have been widely used as statistical emulators for those numerical models. However, since CNNs operate on regular grids, they cannot represent the refined meshes and computational efficiency of finite-element numerical models. Therefore, instead of CNNs, this study adopts an equivariant graph convolutional network (EGCN) as an emulator for the ice sheet dynamics modeling. EGCN reproduces ice thickness and velocity changes in the Helheim Glacier, Greenland, and Pine Island Glacier, Antarctica, with 260 times and 44 times faster computation time, respectively. Compared to the traditional CNN and graph convolutional network, EGCN shows outstanding accuracy in thickness prediction near fast ice streams by preserving the equivariance to the translation and rotation of graphs.
\end{abstract}


\section{Introduction}\label{introduction}

As the mass losses in Greenland and Antarctic ice sheets have accelerated in recent years \cite{Otosaka2023}, it becomes more important to accurately model the ice sheets' behavior and their interaction with climate change. Although several numerical models have been proposed based on the physical understanding of ice flow dynamics (e.g., Stokes equations), those numerical models require intensive computation to solve complex partial differential equations (PDEs). In order to emulate such computationally expensive numerical models, computationally cheap machine learning models have been developed leveraging the parallel processing capability of graphic processing units (GPUs). 


Most of them relied on convolutional neural network (CNN) architectures \cite{Jouvet2022, jouvet2023} that are only compatible with regular Euclidean or grid-like structures, such as images. However, given that finite-element numerical models are implemented on unstructured meshes rather than regular grids, CNN cannot fully represent unstructured meshes of numerical ice sheet models. Therefore, instead of traditional CNN architecture, we choose graph neural network (GNN) as the backbone deep learning architecture to emulate finite-element ice sheet models. Unlike CNN, GNN can operate on any irregular non-Euclidean graph structures (i.e., any data structures with nodes and edges) by updating node features iteratively through message-passing processes between neighboring nodes \cite{zhang2019}. In particular, considering the dynamical behavior of ice sheets, we adopt a special graph convolutional network (GCN) architecture designed to preserve equivariance to rotation and translations of dynamic systems: so-called equivariant graph convolutional network (EGCN) \cite{Satorras2021_EGCN}. 

\begin{figure}[H]
\vskip 0.2in
\begin{center}
\centerline{\includegraphics[width=0.8\columnwidth]{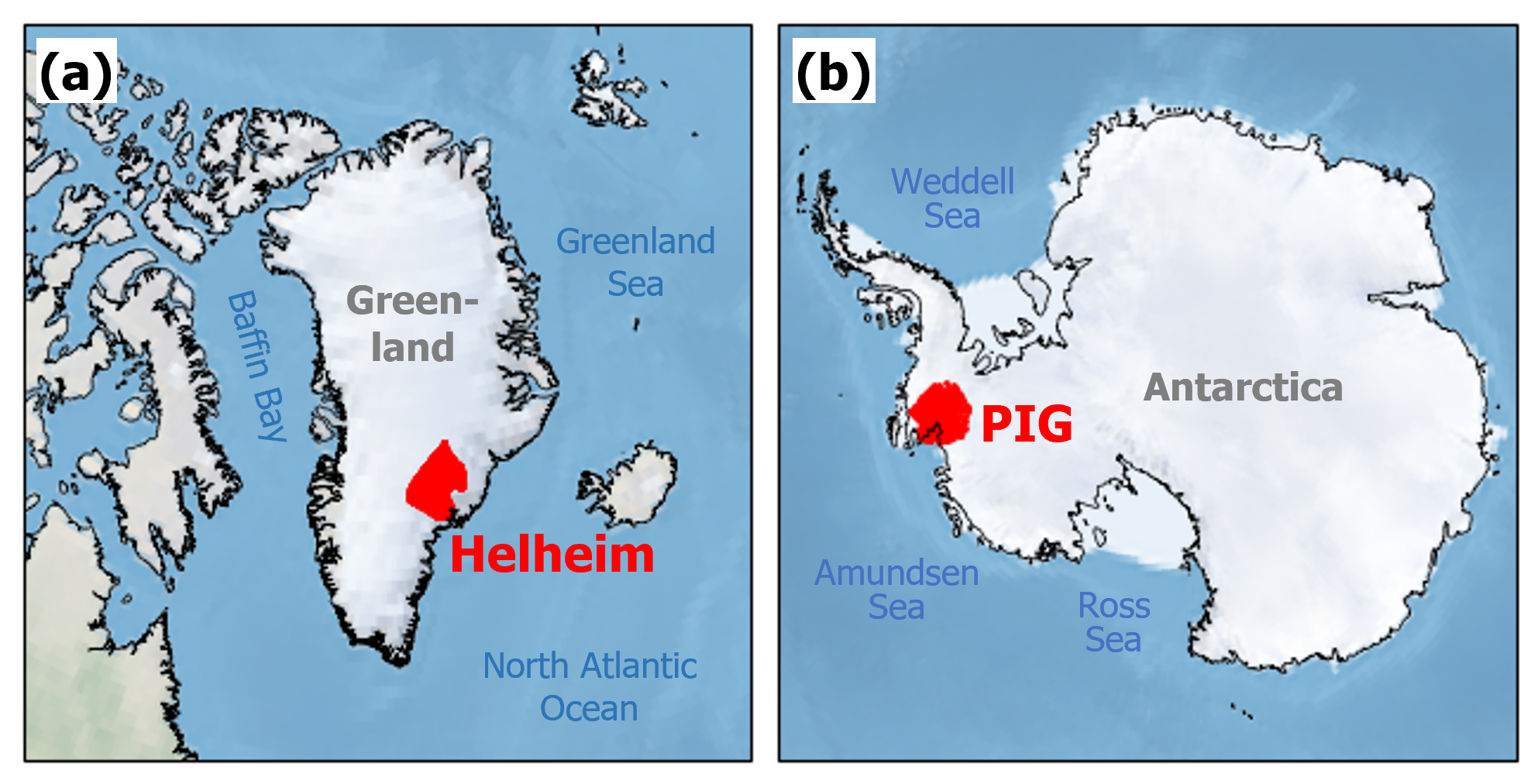}}
\caption{(a) Location of the Helheim Glacier in Greenland; (b) Location of the Pine Island Glacier (PIG) in Antarctica.}
\label{Helheim_PIG}
\end{center}
\vskip -0.2in
\end{figure}

In this study, we aim to develop GNN emulators, particularly EGCN, for the Ice-sheet Sea-level System Model (ISSM) \cite{Larour2012}. The computational efficiency of the ISSM numerical model results from adaptive mesh refinement (AMR), which allocates computational resources depending on the expected precision of ice velocity for individual finite elements \citep{Larour2012, Santos2021}. We take the Helheim Glacier, Greenland (Fig. \ref{Helheim_PIG}a), and Pine Island Glacier (PIG), Antarctica (Fig. \ref{Helheim_PIG}b), as our testing sites because they are the representative ice sheets that have experienced rapid acceleration and mass loss. Given that mass loss of Helheim Glacier is primarily driven by calving \cite{Choi2018_calving_laws, cheng2022helheim} and PIG is driven by basal melting \cite{Joughin2021, jacobs2011}, we predict how the ice sheet dynamics would change by calving and melting parameters.


We train GNN models using the simulation data acquired from the ISSM numerical model and assess their fidelity and computational efficiency for modeling ice dynamics in the Helheim Glacier and PIG. The main contributions of this research are the following:

\begin{itemize}\setlength\itemsep{-0.1em}
\item We adopt EGCN as a statistical emulator to reproduce the ice dynamics simulated from ISSM. This study is the first application of EGCN on ice sheet modeling.
\item We conduct extensive experiments for the Helheim Glacier and PIG to evaluate the potential and superiority of EGCN architecture over traditional GCN and CNN architectures in predicting ice dynamics.
\end{itemize}
\vskip -0.3in



\section{Methodology}\label{method}

The unstructured meshes of ISSM can be regarded as graph structures where the connectivity between nodes is expressed via adjacency matrices. Additionally, since the ice movements change the spatial domains of ice sheets, modeling of ice sheet dynamics can be regarded as a dynamic graph problem. Hence, based on the graph representation of the ISSM meshes, we adopt and test the EGCN, which is specialized for the modeling of dynamically changing graph structures \cite{Satorras2021_EGCN}. We also test two baseline deep learning models: normal GCN and CNN. 

\subsection{Equivariant Graph Convolutional Network (EGCN)}

\begin{figure}[H]
\begin{center}
\centerline{\includegraphics[width=0.9\columnwidth]{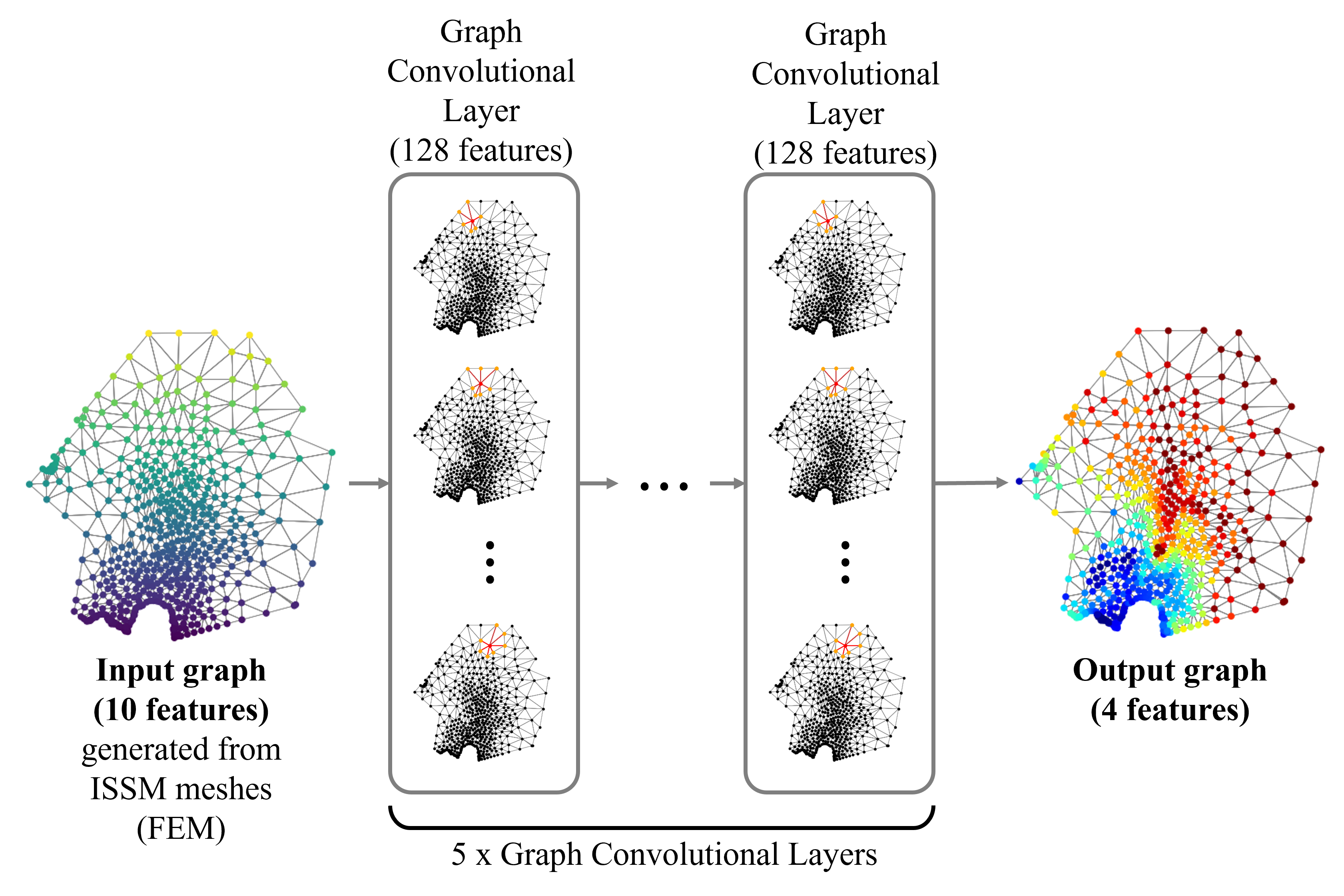}}
\caption{Schematic illustration of GCN architectures.}
\label{gcn_architecture}
\end{center}
\vskip -0.2in
\end{figure}

EGCN is designed to conserve equivariance to rotations, translations, reflections, and permutations in a graph structure \cite{Satorras2021_EGCN}: this EGCN structure has shown greater generalizability to any graph structures of dynamics systems. Let the $l$th graph convolutional layer receives a set of node features $\textbf{h}^{(l)} = \{h_1^{(l)}, h_2^{(l)}, ..., h_N^{(l)}\}$, $h_i^{(l)} \in \mathbb{R}^{F_{l}}$ as the input and produces a new set of node features,  $\textbf{h}^{(l+1)} = \{h_1^{(l+1)}, h_2^{(l+1)}, ..., h_N^{(l+1)}\}$, $h_i^{(l+1)} \in \mathbb{R}^{F_{l+1}}$, for the next $l+1$th layer. $N$ is the number of nodes; $F_{l}$ and $F_{l+1}$ is the number of features in each node at $l$th layer and $l+1$ layer, respectively. Then, an equivariant graph convolutional layer updates node features by using the following equations:

\begin{equation}
m_{ij}=\phi_e(h_i^{(l)}, h_j^{(l)}, ||x_i^{(l)}-x_j^{(l)}||^2, a_{ij})
\label{egcn1}
\end{equation}
\begin{equation}
x_i^{(l+1)}=x_i^{(l)}+C\sum_{j\neq i}(x_i^{(l)}-x_j^{(l)})\phi_x(m_{ij})
\label{egcn2}
\end{equation}
\begin{equation}
m_{i}=x_i^{(l)}+C\sum_{j\neq i}m_{ij}
\label{egcn3}
\end{equation}
\begin{equation}
h_i^{(l+1)}=\phi_h(h_i^{(l)}, m_{i})
\label{egcn4}
\end{equation}
where $a_{ij}$ is the edge attributes, $x_i$ and $x_j$ are the coordinate embeddings for node $i$ and $j$, respectively, and $C$ is a constant for normalization computed as $1/|\mathcal{N}(i)|$. $\mathcal{N}(i)$ is the set of neighbors of node $i$. For the edge attributes, we extract five attributes from the connecting nodes: distance, surface slope, base slope, acceleration of x-component velocity, and acceleration of y-component velocity. $\phi_e$, $\phi_x$, and $\phi_h$ are the edge, position, and node operations, respectively, which are approximated by single-layer MLPs with 128 hidden features. Herein, we regard the x- and y-component ice velocities as the displacement causing the coordinate changes of graphs (i.e., $x_i$ and $x_j$), and ice thickness is represented by hidden features (i.e., $h_i^{(l)}$). Thus, the ice thickness is equivariant to the displacement caused by ice flow. Our EGCN consists of one input layer, five equivariant graph convolutional layers, and one output layer (Fig. \ref{gcn_architecture}). 

\subsection{Graph Convolutional Network (GCN)}

We compare the EGCN with a normal GCN architecture proposed by \citet{Kipf2016_GCN}. Our GCN consists of one input layer, five graph convolutional hidden layers, and one output layer (Fig. \ref{gcn_architecture}). For each graph convolutional layer, the number of features is set to 128. The weights of graph convolutional layers are updated via the layer-wise propagation rule as follows:
\begin{equation}
\label{eq_gcn}
\begin{split}
    & h_i^{(l+1)}=\sigma(\sum_{j\in\mathcal{N}(i)}\frac{1}{c_{ij}}\textbf{W}^{(l)}h_j^{(l)}) \\
\end{split}
\end{equation}
where $c_{ij}$ is the product of the square root of node degrees (i.e., $c_{ij}=\sqrt{|\mathcal{N}(j|)}\sqrt{|\mathcal{N}(i)|}$), $\textbf{W}^{(l)}$ is a layer-specific trainable weight matrix ($\textbf{W}^{(l)} \in \mathbb{R}^{F_{l+1} \times F_{l}}$), and $\sigma(\cdot)$ is an activation function; we use the Leaky ReLU activation function of 0.01 negative slope in this study.

\subsection{Convolutional Neural Network (CNN)}

As a baseline model to be compared with GNN emulators, we train and test a fully convolutional network (FCN), which has a similar architecture to \citet{Jouvet2022}. In the FCN architecture, the hidden graph convolutional layers are replaced with convolutional layers; it consists of five hidden convolutional layers. Each convolution has a 3$\times$3 kernel size and 128 features. The leaky ReLU activation function of 0.01 negative slope is applied after each convolutional layer. Since the FCN takes regular grids as the input and output, we interpolate all the irregular mesh construction of the ISSM into a 1 km regular grid using bilinear interpolation.

\section{Experiment for the Helheim Glacier}\label{results}

In this study, we apply the EGCN, GCN, and FCN emulators to predict the ice sheet dynamics in the Helheim Glacier varying by calving threshold parameter ($\sigma_{\text{max}}$) of the von Mises (VM) calving law \cite{Morlighem2016_calving}. 

\subsection{Preparation of training data}

To generate training datasets for the GNN emulators, we run transient simulations of ice dynamics in the Helheim Glacier between 2007 and 2020. We use the Shelfy-Stream Approximation (SSA) \cite{MacAyeal1989} to explain ice flow in the Helheim Glacier \cite{cheng2022helheim, Choi2018_calving_laws}. For the initial condition of the model, we use the surface velocities from satellite observations \cite{mouginot2017, Mouginot2019}, bed topography from BedMachine Greenland v6 \cite{Morlighem2017}, surface mass balance (SMB) from the Regional Atmosphere Model \cite{tedesco2020unprecedented}, and the ocean thermal forcing from the reanalysis data \cite{Wood2021}. To examine the sensitivity of ice dynamics to the calving parameter $\sigma_{\text{max}}$, we implement the transient solutions for 7 different $\sigma_{\text{max}}$ values (i.e., 0.70, 0.75, 0.80, 0.85, 0.90, 0.95, 1.0 MPa) based on the values proposed by \citet{Choi2018_calving_laws}.

We convert the finite-element simulation results into graph structures by extracting adjacent matrices that represent the connectivity between nodes: for a triangular mesh composed of three nodes, these nodes are connected to each other. Our GNN takes 10 features of graph nodes as inputs, including $\sigma_{\text{max}}$, time, SMB, initial x- and y-component ice velocities, initial surface elevation, bed elevation, initial ice thickness, and initial ice mask. Then, the output layer predicts 4 features of the graph nodes: x-component ice velocity, y-component ice velocity, ice thickness, and ice mask. All those input and output features are normalized into [-1, 1] for stable learning using the nominal maximum and minimum values of those variables.

A total of 1,827 graphs are divided into the train, validation, and test datasets based on the $\sigma_{\text{max}}$ values to assess if our emulator can be generalized for out-of-sample $\sigma_{\text{max}}$ values. The data with $\sigma_{\text{max}}$ of 0.70, 0.80, 0.85, 0.90, and 1.0 MPa are used for training and validation: we randomly split them into 70 \% and 30 \% for training and testing, respectively. The rest of the data with $\sigma_{\text{max}}$ 0.75 and 0.95 MPa are test datasets. Consequently, the number of training, validation, and test datasets is 913, 392, and 522, respectively. The model is optimized by Adam stochastic gradient descent algorithm with the mean square error (MSE) loss function, 400 epochs, and 0.001 learning rate. 


\subsection{Results}



\begin{table}[t]
\caption{Test accuracy of ice velocity and thickness for deep learning emulators for the Helheim Glacier. All matrices are averaged for 2 testing $\sigma_{\text{max}}$ values: 75 and 95 MPa.}
\label{table_accuracy}
\begin{center}
\begin{tabular}{ccccc}
\hline
\multirow{2}{*}{Model} &
\multicolumn{2}{c}{Ice velocity} &
\multicolumn{2}{c}{Ice thickness}\\
& RMSE (m/year) &R & RMSE (m) &R\\
\hline
EGCN & \textbf{95.66} & \textbf{0.990} & \textbf{27.55} & \textbf{0.999}\\
GCN & 104.65 & 0.989 & 48.22 & 0.998\\
FCN & 168.33 & 0.972 & 64.32 & 0.996\\
\hline
\end{tabular}
\end{center}
\vskip -0.2in
\end{table}%

The overall test accuracy of GCN, EGCN, and FCN is shown in Table \ref{table_accuracy}. All deep-learning emulators show outstanding performances in ice velocity and thickness prediction with R-values greater than 0.990 in most cases. EGCN shows the best accuracy among them, with around 10 m/year of lower ice velocity RMSE and 20 m of lower ice thickness RMSE than normal GCN. Whereas GCN uses the spatial distance of neighboring nodes in determining their relative weights in the propagation process (Eq. \ref{eq_gcn}), EGCN uses the message passing from all nodes to preserve the equivariance of the entire graph. The equivariance architecture to the graph rotation and translation could guarantee more generalizability to any graph conditions, leading to the improvement of model fidelity. In particular, EGCN can potentially represent the ice thickness changes induced by ice velocity because the spatial embeddings are used to predict hidden feature embeddings (Eq. \ref{egcn1}-\ref{egcn4}). Additionally, we highlight that GNN emulators outperform traditional FCN. Since FCN uses the interpolated 1-km regular grid for all points without any adjustment to ice velocity, FCN inhibits delineating boundary conditions in a fine resolution, leading to significant errors. Since ISSM and GNN emulators use irregular meshes, which assign a finer resolution for fast ice and a coarser resolution for static ice, they can describe the ice boundaries more accurately with fine resolutions.

\section{Experiments for the Pine Island Glacier}\label{pig}

We applied the same GCN, EGCN, and FCN architectures to the PIG in Antarctica. Although the same input and output features are used for the PIG emulator, only the calving parameter $\sigma_{\text{{max}}}$ is replaced with basal melting rate because the ice dynamics in PIG are likely driven by melt-driven thinning near the grounding line rather than calving.

\subsection{Preparation of training data}

Similar to the Helheim experiment, the training datasets are also collected from the ISSM transient simulation for 20 years using the SSA. We implement the ISSM simulations for different annual basal melting rate scenarios ranging from 0 to 70 m/year for every 2 m/year. Consequently, we collect 8,640 graphs from the experiments with 36 different melting rates. We divide the 8,640 graphs into train, validation, and test datasets based on the melting rate values: melting rates of 10, 30, 50, and 70 m/year are used as validation datasets, melting rates of 0, 20, 40, and 60 m/year as test datasets and the rest are used as training datasets. The total graphs for training, validation, and testing are 6,720, 960, and 960, respectively. We use the following observation data as the initial ice conditions: ice velocity from the NASA Making Earth System Data Records for Use in Research Environments (MEaSUREs) \cite{MEASURE_velocity}, 1 km Antarctic digital elevation model (DEM) \cite{Bamber2009_DEM}, bedrock topography data of the Amundsen Sea continental shelf \cite{Nitsche2007_bedrock}, Antarctic surface temperature data \cite{Comiso2000_temperature}, and Antarctic surface mass balance data \cite{Vaughan1999_SMB}.

\subsection{Results}

Table \ref{table_PIG_fidelity} summarizes the accuracy of ice velocity and thickness prediction from deep learning emulators. All deep learning emulators show significant performance in both ice velocity and thickness predictions with R greater than 0.997. GNN emulators, including EGCN and GCN, show better accuracy than FCN in ice velocity prediction, and EGCN shows the best accuracy. EGCN shows a remarkably low ice thickness RMSE compared to the other emulators, which agrees with the results from the Helheim Glacier. Therefore, the equivariant architecture of EGCN helps improve the predictability of ice thickness changes caused by ice dynamics in the PIG as well.

\begin{table}[h]
\centering
\caption{Test accuracy of ice velocity and thickness for deep learning emulators for the PIG. All matrices are averaged for 4 testing melting rates: 0, 20, 40, and 60 m/year.}
\label{table_PIG_fidelity}
\begin{tabular}[t]{ccccc}
\hline
\multirow{2}{*}{Model} &
\multicolumn{2}{c}{Ice velocity} &
\multicolumn{2}{c}{Ice thickness}\\
&RMSE (m/year) &R &RMSE (m) &R \\
\hline
EGCN &\textbf{55.29} &\textbf{0.997} &\textbf{14.75} &\textbf{0.999} \\
GCN &55.99 &0.997 &34.61 &0.999 \\
FCN &56.69 &0.997 &20.18 &0.999 \\
\hline
\end{tabular}
\end{table}

\section{Computational performance}\label{discussion}

We record and compare the time to generate the final transient simulations for the Helheim Glacier and PIG experiments (Table \ref{table_time}). The computation time of ISSM is the total elapsed time spent on a single node of the Texas Advanced Computing Center (TACC) Frontera supercomputing cluster, which is equipped with 56 cores of Intel 8280 Cascade Lake CPUs (192 GB memory). The computation times of deep learning emulators are the total elapsed time on a CPU (Intel(R) Core(TM) i7-11700F; 32 GB memory) and GPU (NVIDIA GeForce RTX 3070; 24GB memory). In the Helheim experiment, GCN shows the most dramatic speed-up by around 560 times faster computation time than ISSM, followed by EGCN (260 times speed-up) and FCN (250 times speed-up). In the PIG experiment, GCN and EGCN show 50 times and 44 times faster computation time than ISSM, respectively. FCN takes the longest time to reproduce the 20-year ice sheet dynamics, with only 21 times speed-up compared to the ISSM simulation.

\begin{table}
\begin{center}
\caption{Total computational time (in seconds) for producing ice sheet 20-year transient simulations in the PIG for 36 melting rates and training time to train deep learning emulators.}
\label{table_time}
\begin{tabular}[t]{c|cc|cc}
\hline
\multirow{2}{*}{Model} & \multicolumn{2}{c|}{Helheim} & \multicolumn{2}{c}{PIG}\\
\cline{2-5}
& CPU & GPU & CPU & GPU \\
\hline
ISSM & 6681.25 & - &712.96 & - \\
GCN & \textbf{200.58} & \textbf{11.89} & \textbf{125.05} & \textbf{14.18} \\
EGCN & 759.37 & 25.42 & 607.77 & 16.14 \\
FCN & 851.71 & 27.01 & 1932.00 & 33.34 \\
\hline
\end{tabular}
\end{center}
\vskip -0.2in
\end{table}

\section{Conclusions}

This study introduces an equivariant graph convolutional network (EGCN) as a deep learning emulator to reproduce ice dynamics modeled by the Ice-sheet and Sea-level System Model (ISSM) and compares its performance with traditional graph convolutional network (GCN) and convolutional neural network (CNN). When the EGCN is trained with transient simulations in the Helheim Glacier, Greenland, and Pine Island Glacier (PIG), Antarctica, EGCN shows better accuracy than the others in predicting ice velocity and thickness by preserving the equivariance of graph structures. By reducing computational time 50-500 times compared to the CPU-based numerical models, the EGCN and GCN emulators will be promising tools to investigate the impacts of parameterization on future ice behavior, which will contribute to the improvement of the prediction accuracy of ice sheet mass loss and sea level rise.








\bibliography{reference}
\bibliographystyle{icml2024}

\newpage
\appendix
\onecolumn

\end{document}